\documentclass[review]{elsarticle}

\usepackage{amsmath,amsfonts}
\usepackage{bm}

\usepackage{times}
\usepackage{epsfig}
\usepackage{amsmath}
\usepackage{amsfonts}
\usepackage{amssymb}
\usepackage{url}
\usepackage{latexsym}
\usepackage{subfigure}
\usepackage{algorithm2e}
\usepackage{caption}
\usepackage{subfigure}

\usepackage{framed,multirow}

\usepackage{chngpage}

\begin{document}

\begin{frontmatter}

\title{An empirical study of automatic wildlife detection using drone thermal imaging and object detection}

\author[a]{Miao Chang}
\ead{changmiao@deakin.edu.au}

\author[a]{Tan Vuong}
\ead{vuongt@deakin.edu.au}

\author[a]{Manas Palaparthi}
\ead{m.palaparthi@deakin.edu.au}

\author[b]{Lachlan Howell}
\ead{l.howell@deakin.edu.au}

\author[a]{Alessio Bonti}
\ead{a.bonti@deakin.edu.au}

\author[c]{Mohamed Abdelrazek}
\ead{mohamed.abdelrazek@deakin.edu.au}

\author[a]{Duc Thanh Nguyen\corref{d}}
\ead{duc.nguyen@deakin.edu.au}

\address[a]{Deakin University,\\
 				 School of Information Technology, Geelong VIC 3220, Australia}

\address[b]{Deakin University,\\
                School of Life and Environmental Sciences, Geelong VIC 3220, Australia}
                
\address[c]{Deakin University,\\
                Applied Artificial Intelligence Institute, Burwood VIC 3125, Australia}

\cortext[d]{Corresponding Author}

\begin{abstract}
Artificial intelligence has the potential to make valuable contributions to wildlife management through cost-effective methods for the collection and interpretation of wildlife data. Recent advances in remotely piloted aircraft systems (RPAS or ``drones'') and thermal imaging technology have created new approaches to collect wildlife data. These emerging technologies could provide promising alternatives to standard labourious field techniques as well as cover much larger areas. In this study, we conduct a comprehensive review and empirical study of drone-based wildlife detection. Specifically, we collect a realistic dataset of drone-derived wildlife thermal detections. Wildlife detections, including arboreal (for instance, koalas, phascolarctos cinereus) and ground dwelling species in our collected data are annotated via bounding boxes by experts. We then benchmark state-of-the-art object detection algorithms on our collected dataset. We use these experimental results to identify issues and discuss future directions in automatic animal monitoring using drones.

\end{abstract}


\end{frontmatter}


\section{Introduction}
\label{sec:introduction}

Wildlife, ecosystem and habitat protection require monitoring of wildlife species where regular and reliable population assessments and census trends are necessary~\cite{CSIRO}. Conventional ground-based survey techniques are time-consuming and labour-intensive. Demand for automatic surveillance systems continues to grow, supported by emerging technologies that have greatly improved the power, accuracy and efficiency of ecological data collection.

Artificial intelligence (AI) research has been used to develop automated wildlife monitoring systems. For instance, research in audio signal processing and machine learning are used to detect and classify bird calls~\cite{DBLP:conf/icassp/NealBRF11, Nguyen_2017, DBLP:journals/ecoi/MaegawaUSTKHM21, DBLP:conf/icassp/DentonWH22}. Image recognition techniques are adopted to detect and identify animals from camera traps~\cite{DBLP:journals/ejivp/YuWKJWH13, 7523423, Chandrakar_2022}. Computer vision methods are applied to recognise aquatic creatures captured by autonomous underwater vehicles~\cite{Chen_2020, Merencilla2021SharkEYEAD, DBLP:journals/mta/WeiYTFN21, 9261357, DBLP:journals/tnn/YehLKHLCW22}. Recently, drones equipped with cameras and automated detection technologies have been applied to wildlife monitoring~\cite{Bennitt_2019, Corcoran_2021}. Drones offer many advantages including large-scale data collection, non-invasive and real-time monitoring, cloud storage, and fast playback~\cite{DBLP:journals/remotesensing/WuWLZJWCGLSLLJ22}. In addition, thermal imaging devices when attached to drones, can provide further aid to detecting cryptic animals and distinguishing species from their surrounding environment in low-lighting conditions (e.g., night time)~\cite{Mirka_2022, DBLP:journals/remotesensing/LeeSK21}.

While the increasing volume of wildlife data is useful to ecological studies, it raises a demand on effective methods to process and analyse large-scale datasets. Manual animal counting can be labour-intensive and prone to interpreter bias, resulting in less reliable estimates~\cite{Corcoran_2019, DBLP:journals/sensors/HongHKLK19}. For instance, without automated detection capabilities, aerial imagery is not perceived as a viable option for kangaroo monitoring in Australia due to the intensive effort required to process high-volume data~\cite{DBLP:journals/remotesensing/UlhaqACKLP21}. Automatic counting, on the other hand, can quickly process large numbers of images through AI-based methods, reducing deviations, saving time and labour, and improving accuracy. Recent advances in computing hardware and resources have made AI algorithms feasible in many real-world applications.

A commonly adopted approach to automate the counting process is to apply object detection algorithms~\cite{DBLP:conf/iccv/LinGGHD17, DBLP:journals/pami/RenHG017, DBLP:conf/cvpr/RedmonF17}, examples of AI-based methods, to find animal individuals from recorded aerial imagery. Object detection aims to localise instances of objects of interest, e.g., animals, and specify the semantic class, e.g., species, for each detection~\cite{Zhao2019ObjectDW}. As shown in the  literature~\cite{Corcoran_2019, DBLP:journals/remotesensing/UlhaqACKLP21, DBLP:journals/ecoi/PetsoJMBM21, DBLP:journals/remotesensing/WinsenDCH22, f4e95cd417724949b508a04311899514}, object detection has proven promise to enable automatic animal counting solutions, solving issues of laborious processing of large-scale wildlife datasets and aiding wildlife monitoring efforts. In addition, the availability of open-source object detectors has made object detection-based solutions feasible.

Given the rising uptake of drones and thermal imaging in automatic wildlife monitoring, we aim to conduct an empirical study of drone-based wildlife monitoring through object detection. Our work differs from other studies, e.g.,~\cite{Corcoran_2019, DBLP:journals/sensors/HongHKLK19, Corcoran_2021, DBLP:journals/remotesensing/WinsenDCH22, Petso_2022}, in several aspects. First, the novelty of our work relies on the empirical aspect of our review where we evaluated existing wildlife monitoring methods empirically, rather than just describing the methods as in other surveys, e.g.,~\cite{Corcoran_2021, Petso_2022}. Second, compared with~\cite{Corcoran_2019, DBLP:journals/sensors/HongHKLK19, DBLP:journals/remotesensing/WinsenDCH22}, we benchmark more modern object detection techniques. Third, since we evaluate existing methods on the same dataset using the same settings, observations drawn from our experiments would be more conclusive. In summary, we make the following contributions in our work.
\begin{itemize}
    \item A concise review of drone-based wildlife monitoring research since 2018 (Section~\ref{sec:related_work});
    \item An annotated dataset of wildlife thermal detections captured using drones in forested environments (Section~\ref{sec:dataset});
    \item Extensive experiments and benchmark results of many modern state-of-the-art object detection algorithms in detecting thermal signatures (Section~\ref{sec:benchmark}).
\end{itemize}

\section{Literature review}
\label{sec:related_work}

\subsection{Drone types and settings}

Depending on the species-specific or habitat-specific characteristics of the studied species, different settings can be applied to the drone. For instance, to collect koala imagery, Winsen et al.~\cite{DBLP:journals/remotesensing/WinsenDCH22} set the drone at an altitude of 60 m. This setting was subject to the height of the canopy and the terrain of the site, and therefore was adjusted to maintain a flight altitude about 30 m above the canopy. In~\cite{DBLP:journals/sensors/HongHKLK19}, the flight altitude was set at 100 m to detect waterbird species and the drone captured images in a high resolution to minimise the long-distance temperature measurement error. Cox et al.~\cite{Cox_2021} evaluated different drone thermal sensors for detecting rabbit burrows and suggested a minimum export rate of 30 Hz to ensure image quality.

Compared with ground surveys, drone-based surveys can collect wildlife data in severe conditions such as night time, winter, due to thermal sensors use. In several cases, both RGB and thermal information are combined to enhance animal detection~\cite{DBLP:journals/remotesensing/LeeSK21}.We summarise technical parameters of common drone settings in Table~\ref{tab:UAV_setting}.

\begin{table}
    \small
    \centering
    \caption{Summary of drone-based animal monitoring studies since 2018.}
    \begin{tabular}{|l|l|l|l|}
        \hline
        \textbf{Study (year)} & \textbf{Drone \& sensor} & \textbf{Image resolution} & \textbf{Object detector} \\
        \hline
        \cite{KELLENBERGER2018139} (2018) & SenseFly eBee & $4000\times3000$ & CNN \\
        & Canon PowerShot S110 (RGB) & & \\ 
        \hline
        \cite{Corcoran_2019} (2019) & Matrice 600 Pro & $640 \times 512$ & Faster RCNN,\\
        & FLIR Tau 2 (thermal) & & YOLO \\
        \hline
        \cite{DBLP:journals/sensors/HongHKLK19} (2019) & K-mapper & $6480 \times 4320$ & RetinaNet, SSD,\\
        & NX-500 (RGB) & & R-FCN,\\
        & & & Faster RCNN,\\
        & & & YOLO\\
        \hline
        \cite{DBLP:journals/remotesensing/LeeSK21} (2021) & Matrice 210 & $640 \times 512$ (thermal) & Sobel detector \\
        & FLIR Zenmuse XT2 (thermal, RGB) & $4000 \times 3000$ (RGB) & \\
        \hline
        \cite{DBLP:journals/remotesensing/UlhaqACKLP21} (2021) & DJI Inspire-I RPA & $640 \times 512$ & YOLO \\
        & FLIR Zenmuse (thermal) & & \\
        \cline{2-4}
        & DJI Matrice 210 RPA & $640 \times 512$ & YOLO \\
        & FLIR Zenmuse XT 640 (thermal) & & \\
        \hline
        \cite{DBLP:journals/ecoi/PetsoJMBM21} (2021) & DJI Phantom III & $4000 \times 3000$ & SSD, YOLO \\
        & 1/2.3'' CMOS, effect. pixels: 12.4M (RGB) & &\\
        \hline
        \cite{DBLP:journals/remotesensing/WinsenDCH22} (2021) & Matrice 600 Pro & $640 \times 512$ & Faster RCNN,\\
        & FLIR Tau 2 (thermal) & & YOLO\\
        \hline
        \cite{Sudholz_2021} (2021) & DJI Matrice 600 Pro 
        & $640 \times 512$ & Faster RCNN,\\
        & FLIR Tau 2 (thermal) & & YOLO\\
        \hline
        \cite{DBLP:journals/remotesensing/WuWLZJWCGLSLLJ22} (2022) & Dahua & $1024 \times 768$ & Depth density\\
        & ONVIF (RGB) & & network\\
        \hline
        \cite{Mirka_2022} (2022) & Leica Aibot A X20 & $640 \times 512$ (thermal) & Image matching \\ 
        & WIRIS Pro (thermal, RGB) & $1920 \times 1080$ (RGB) & \\
        \hline
        \cite{f4e95cd417724949b508a04311899514} (2022) & DJI Inspire-II & $6016 \times 4008$ & RetinaNet \\ 
        & Zenmuse X7 (RGB) & & \\
        \hline
        Ours (2023) & Mavic 2 Enterprise Advanced & $640 \times 512$ (thermal) & RetinaNet,\\
        & Uncooled VOx Microbolometer (thermal) & $3840 \times 2160$ (RGB) & Faster RCNN,\\
        & 1/2'' CMOS, effect. pixels: 48M (RGB) & & YOLO, FCOS,\\
        & & & DDETR\\
        \hline
    \end{tabular}
    \label{tab:UAV_setting}
\end{table}

\subsection{Animal detection}

After images are collected, object detection algorithms are used to detect animals. Animal detection aims to locate animal species in an input image. The task of animal detection has various challenges. Since drones acquire images from a high altitude, animal signatures become extremely small and many important characteristics (e.g., colour, texture, shape cannot be well captured). Animals also often blend into cluttered backgrounds (e.g., dense vegetation), making them indistinguishable from their surrounding environment.

Commonly used object detectors for animal detection from drone-based imagery include RetinaNet~\cite{DBLP:conf/iccv/LinGGHD17}, Faster RCNN~\cite{DBLP:journals/pami/RenHG017}, and YOLO~\cite{DBLP:conf/cvpr/RedmonF17}. These detectors are network architectures divided into two branches to perform two sub-tasks: bounding box estimation (for object localisation) and bounding box classification (for object identification). When adapting these tools to animal detection, one needs to customise the last layer in the classification branch of the detector with new neurons corresponding to animal classes of interest, and then re-train the modified detector on an animal dataset. Some other older detectors such as SSD~\cite{10.1007/978-3-319-46448-0_2} and R-FCN~\cite{DBLP:conf/nips/DaiLHS16} were used for bird detection in~\cite{DBLP:journals/sensors/HongHKLK19}.

Several methods create their own model for animal detection. For instance, Wu et al.~\cite{DBLP:journals/remotesensing/WuWLZJWCGLSLLJ22} proposed a depth density estimation network (based on ResNet~\cite{DBLP:conf/cvpr/HeZRS16}) to generate a heat map of waterbirds density. Kellenberger et al.~\cite{KELLENBERGER2018139} and Barbedo et al.~\cite{s20072126} divided an input image into a regular grid, on which a CNN was built to estimate the probability of having animals on every grid cell.

A common practice is to adopt a model pre-trained on large-scale datasets, e.g., ImageNet~\cite{DBLP:conf/cvpr/DengDSLL009}, COCO~\cite{DBLP:conf/eccv/LinMBHPRDZ14}, then fine-tune it on a target domain~\cite{Corcoran_2021}. Since each detector has its own advantages, to maximise the benefit brought by multiple detectors, one can combine different detectors via fusing their detection results~\cite{Corcoran_2019, Sudholz_2021} or through ensemble learning~\cite{DBLP:journals/remotesensing/WinsenDCH22}. Corcoran et al.~\cite{Corcoran_2019} applied temporal information to perform animal tracking to remove false alarms. Ulhaq et al.~\cite{DBLP:journals/remotesensing/UlhaqACKLP21} replaced the upsampling operators in the YOLO’s architecture with dilated convolutions to address feature corruption in the upsampling steps. Several strategies were proposed in~\cite{KELLENBERGER2018139} to handle class imbalance issue in training sets, including using class distribution in a training set to weight the loss function and cropping training images to reduce background bias.

Traditional image processing techniques also exist for animal detection. Lee et al.~\cite{DBLP:journals/remotesensing/LeeSK21} applied the Sobel edge detector to extract edges from input images, then verified important edges using thermal information, requiring a lot of user-defined parameters. Mirka et al.~\cite{Mirka_2022} detected monkeys by matching moving objects between frames using ORB~\cite{DBLP:conf/iccv/RubleeRKB11} (a handcrafted image descriptor like SIFT~\cite{DBLP:journals/ijcv/Lowe04}). We present object detectors used in existing studies in Table~\ref{tab:UAV_setting}.

\section{Dataset}
\label{sec:dataset}

\subsection{Field method and data collection}

We deployed a Mavic 2 Enterprise Advanced equipped with a $640 \times 512$-pixel thermal camera ($\sim$9 mm focal length), and a $3840 \times 2160$-pixel RGB camera (see Table~\ref{tab:UAV_setting}). Both sensors worked at 30 Hz. The white-hot thermal pallet was used, and the thermal sensor was set to high gain.

We focused on developing a real-world dataset of all suspected wildlife thermal signatures detected in a given study area. We surveyed 25 ha plots of native forest in three locations in East Gippsland, Victoria, Australia, including two plots in Gelantipy (37$^{\circ}$13'31.65''S, 148$^{\circ}$15'34.34''E and 37$^{\circ}$16'53.30''S, 148$^{\circ}$14'39.24''E) and two plots in Orbost (37$^{\circ}$44'26.88''S, 148$^{\circ}$12'15.21''E and 37$^{\circ}$43'59.75''S, 148$^{\circ}$12'16.41''E) across 17-20 May and 6-8 June 2022 in Gelantipy, and across 20-23 June 2022 in Orbost. Sites in Gelantipy consisted of Montane Grassy Woodland (composed of Broad-leaved peppermint, Eucalyptus dives, and Candlebark, Eucalyptus rubida) and Valley Grassy Forest (composed of River peppermint, Eucalyptus elata, Manna gum, Eucalyptus viminalis, White stringybark, Eucalyptus globoidea, Red box, Eucalyptus polyanthemos). Sites in Orbost consisted of Lowland forest with dominant tree species Silvertop Ash (Eucalyptus sieberi). 

Minimum temperatures for the survey period ranged from 0.3–8$^{\circ}$C, and maximum temperatures ranged from 6.1–18.4$^{\circ}$C across the survey period. Wind speed was consistently $<$7.7 m/s. Surveys occurred between $\sim$6pm–2am to maximise temperature differential and contrast between animals and the surrounding environment. Flights were programmed using UgCS 4.9.814 and were flown in a lawn mower pattern composed of parallel linear survey lines with 10\% overlap. Flight altitude was 50-70 m above ground level depending on site-specific terrain and tree height. Flight speed was 5-6 m/s. Since the data were captured at night, only thermal images were retained and used in our experiments.

Surveys were conducted under Deakin University’s Animal Ethics Committee Wildlife-Burwood protocol: B08-2022.

\subsection{Data annotation}

After collecting imagery data, we manually extracted images containing suspected thermal detections of wildlife species. We used CVAT\footnote{\url{https://cvat.org}}, a free and open-source annotation tool to label bounding boxes of suspected wildlife thermal signatures in the extracted images. During surveys in-field, we were only able to validate a small number of detected thermal signatures (64 signatures) using on-ground observers due to weather and other external factors. Due to a lack of thermal signature validation, we annotated only one class: ``suspected wildlife signature''. We note that this limitation does not compromise the aim of this empirical study, benchmarking available object detection models on their detection accuracy and efficiency. 

\begin{figure}
    \centering
    \includegraphics[width=\columnwidth]{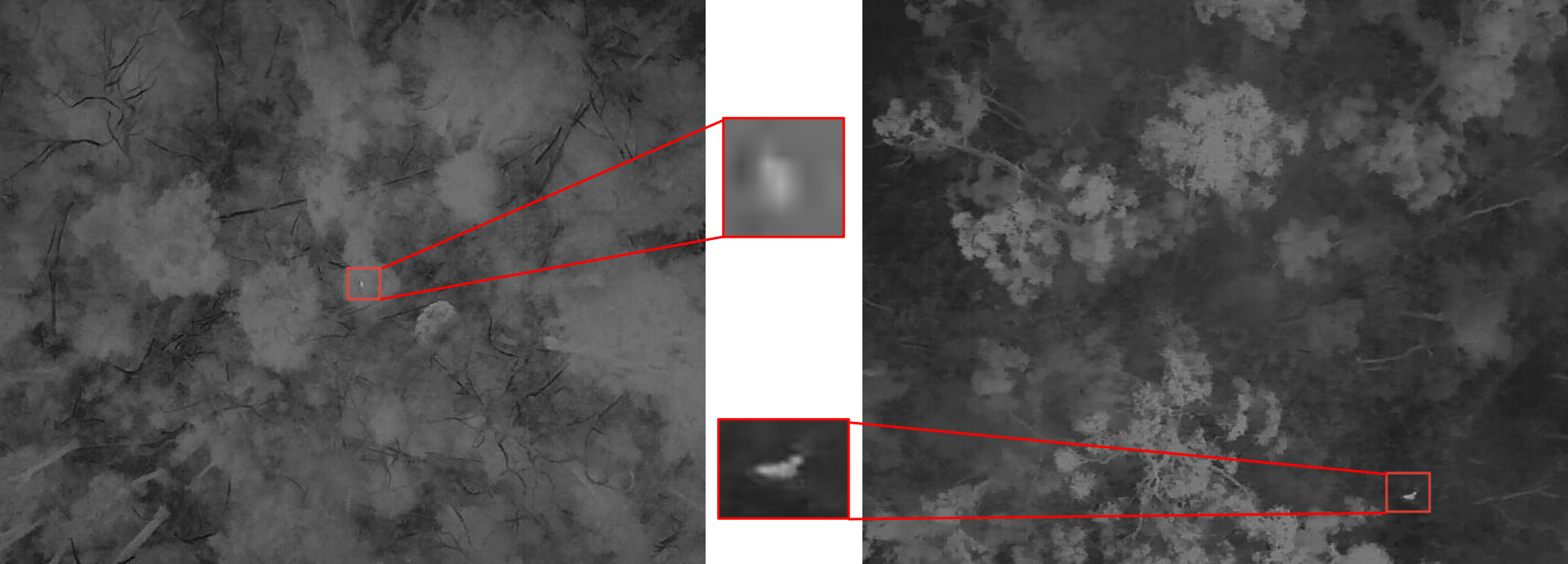}
    \caption{Illustration of annotated animals (with close-ups).}
    \label{fig:annotation}
\end{figure}

Annotated thermal signatures are very small ($32 \times 32$ pixels relative to the image resolution of $640 \times 512$ pixels). Annotation resulted in a dataset of 1,210 images with 1,814 annotated thermal signatures. We split the dataset into two subsets: training (850 images with 1,273 thermal signatures) and validation (360 images with 541 thermal signatures). We visualise several annotation results in Figure~\ref{fig:annotation}.

\section{Benchmark of animal detection}
\label{sec:benchmark}

\subsection{Baselines}
\label{sec:baselines}

We experimented with the following object detection baselines: RetinaNet~\cite{DBLP:conf/iccv/LinGGHD17}, Faster RCNN~\cite{DBLP:journals/pami/RenHG017}, YOLO~\cite{DBLP:conf/cvpr/RedmonF17}, FCOS~\cite{DBLP:conf/iccv/TianSCH19}, and DDETR~\cite{DBLP:conf/iclr/ZhuSLLWD21}. These were selected due to their high detection rate and fast computational speed, proven in many applications and on various datasets. RetinaNet, YOLO, and Faster RCNN make use of anchor boxes in finding object bounding boxes. In contrast, FCOS and DDETR are anchor box-free methods. DDETR is built based on Vision Transformer~\cite{dosovitskiy2021an}, a modern network architecture.

Some detectors have different variants. For instance, YOLO has a long history of development with many variants corresponding to different amendments in the architecture (for both feature learning and object location prediction)~\cite{Jiang_2022}. In this paper, we experimented with the versions 4, 5, and 7 of YOLO due to their contemporary as well as proven superiority over other versions in the YOLO family~\cite{Nepal_2022, DBLP:journals/corr/abs-2207-02696}. Moreover, YOLOv5 (the fifth version of YOLO) has also shown promise to detect objects in drone imagery~\cite{Baidya_2022, Nepal_2022}. For YOLOv4, we evaluated the original version in~\cite{DBLP:journals/corr/abs-2004-10934, DBLP:conf/cvpr/WangBL21} and its variants, e.g., with the use of Cross Stage Partial connections~\cite{wang2020cspnet} (csp), with different activation functions (mish, leaky), and network sizes (x: large, s: small). For YOLOv5, we experimented with 4 variants corresponding to 4 different scales (s: small, m: medium, x: large, and x: extra-large). For YOLOv7~\cite{DBLP:journals/corr/abs-2207-02696}, we benchmarked variants developed for different hardware specifications, e.g., edge GPU (YOLOv7-tiny), normal GPU (YOLOv7), cloud GPU (YOLOv7-w6), different sub-structures in the network architecture, e.g., YOLOv7-x, YOLOv7-e6e, YOLOv7-d6, and different backbones, e.g., YOLOv7-PRB, i.e., with the Parallel Residual Bi-Fusion (PRB) Feature Pyramid Network~\cite{DBLP:journals/tip/ChenCHC21}.

For Faster RCNN, we evaluated commonly used backbones including ResNet50, ResNet101~\cite{DBLP:conf/cvpr/HeZRS16} and HRNet~\cite{8953615} (which is designed for high-resolution images). Both the ResNet50 and ResNet101 were also used and evaluated in RetinaNet. These variants have varying numbers of parameters, resulting in different training and inference speed. We summarise the object detectors experimented in our study in Table~\ref{tab:model_summary}. 

\begin{table*}
    \small
    \centering
    \caption{Summary of object detection models regarding to Floating Point Operations Per Second - FLOPS (G), number of parameters (M), training time (hours:minutes), inference speed (frames per second - fps).}
    \begin{tabular}{|l|c|c|c|c|}
        \hline
        \textbf{Model} & \textbf{FLOPS} $\downarrow$ & \textbf{No. Param} $\downarrow$& \textbf{Training time} $\downarrow$ & \textbf{Inference speed} $\uparrow$ \\
        \hline
        RetinaNet (ResNet50) & 81.69 & 36.10 & 1h:32m & 21.20 \\
        \hline
        RetinaNet (ResNet101) & 110.22 & 54.99 & 1h:55m & 17.80 \\
        \hline
        Faster RCNN (ResNet50) & 78.12 & 41.12 & 1h:50m & 18.70 \\
        \hline
        Faster RCNN (ResNet101) & 108.55 & 60.11 & 2h:55m & 24.10 \\
        \hline
        Faster RCNN (HRNet) & 109.46 & 46.87 & 2h:00m & 16.90 \\
        \hline
        YOLOv4 & 107.10 & 63.94 & 1h:10m & 19.90 \\
        \hline
        YOLOv4-csp-s-leaky & 16.40 & 8.06 & 0h:43m & 13.20 \\
        \hline
        YOLOv4-csp-leaky & 107.50 & 52.50 & 0h:53m & 18.20 \\
        \hline
        YOLOv4-csp-x-leaky & 184.40 & 99.22 & 1h:20m & 21.90 \\
        \hline
        YOLOv4-csp-s-mish & 16.40 & 8.05 & 0h:30m & 14.20 \\
        \hline
        YOLOv4-csp-mish & 107.50 & 52.50 & 0h:57m & 18.40 \\
        \hline
        YOLOv4-csp-x-mish & 184.40 & 99.22 & 1h:45m & 21.90 \\
        \hline
        YOLOv5s & 15.90 & 7.02 & 0h:21m & 73.40 \\
        \hline
        YOLOv5m & 48.20 & 20.87 & 0h:28m & 52.42 \\
        \hline
        YOLOv5l & 108.20 & 46.13 & 0h:45m & 40.78 \\
        \hline
        YOLOv5x & 203.80 & 86.17 & 1h:22m & 26.21 \\
        \hline
        YOLOv7 & 105.10 & 37.20 & 1h:17m & 11.90 \\
        \hline
        YOLOv7-tiny & 13.20 & 6.01 & 1h:00m & 3.80 \\
        \hline
        YOLOv7-w6 & 101.80 & 80.90 & 2h:36m & 10.80 \\
        \hline
        YOLOv7-x & 188.90 & 70.82 & 5h:28m & 13.10 \\
        \hline
        YOLOv7-e6e & 225.40 & 164.82 & 3h:25m & 20.40 \\
        \hline
        YOLOv7-d6 & 198.30 & 152.89 & 5h:48m & 17.20 \\
        \hline
        YOLOv7-PRB & 255.30 & 102.15 & 1h:35m & 25.20 \\
        \hline
        FCOS & 109.06 & 50.78 & 1h:20m & 26.40 \\
        \hline
        DDETR & 27.40 & 39.82 & 2h:58m & 16.20 \\
        \hline
    \end{tabular}
    \label{tab:model_summary}
\end{table*}

We adopted publicly released code repositories including Ultralytics\footnote{\url{https://github.com/ultralytics/yolov5}} for YOLOv5, mmdetection\footnote{\url{https://github.com/open-mmlab/mmdetection}} for RetinaNet, Faster RCNN, and FCOS, and the official implementation by the authors of YOLOv4\footnote{\url{https://github.com/WongKinYiu/PyTorch_YOLOv4}}, YOLOv7\footnote{\url{https://github.com/WongKinYiu/yolov7}}, YOLOv7-PRB\footnote{\url{https://github.com/pingyang1117/PRBNet_PyTorch}}, and DDETR\footnote{\url{https://github.com/fundamentalvision/Deformable-DETR}}.

\subsection{Experiments, results, and discussions}

We implemented all the baselines in Pytorch 1.8.0 with Cuda 10.1. All experiments were conducted on 20 cores of Intel(R) Xeon(R) CPU E5-2630 v4@2.20GHz and 8 NVIDIA GeForce GTX 1080 Ti GPUs. Each model and its variants were trained using the same settings (see Table~\ref{tab:setting}).

\begin{table}
    \small
    \centering
    \caption{Experimental settings.}
    \begin{tabular}{|l|c|c|c|c|}
        \hline
        \textbf{Model} & \textbf{Learning rate} & \textbf{Optimiser} & \textbf{Batch size} & \textbf{Epochs} \\
        \hline
        RetinaNet & 0.001 & SGD & 16 & 100 \\
        \hline
        Faster RCNN & 0.02 & SGD & 16 & 100 \\
        \hline
        YOLOv4 & 0.01 & SGD & 8 & 100 \\
        \hline
        YOLOv5 & 0.01 & SGD & 32 & 100 \\
        \hline
        YOLOv7 & 0.01 & SGD & 4 & 100 \\
        \hline
        FCOS & 0.005 & SGD & 16 & 100 \\
        \hline
        DDETR & 0.0002 & Adam & 8 & 100 \\
        \hline
    \end{tabular}
     \label{tab:setting}
\end{table}

\begin{figure}
    \centering
    \includegraphics[width=\columnwidth]{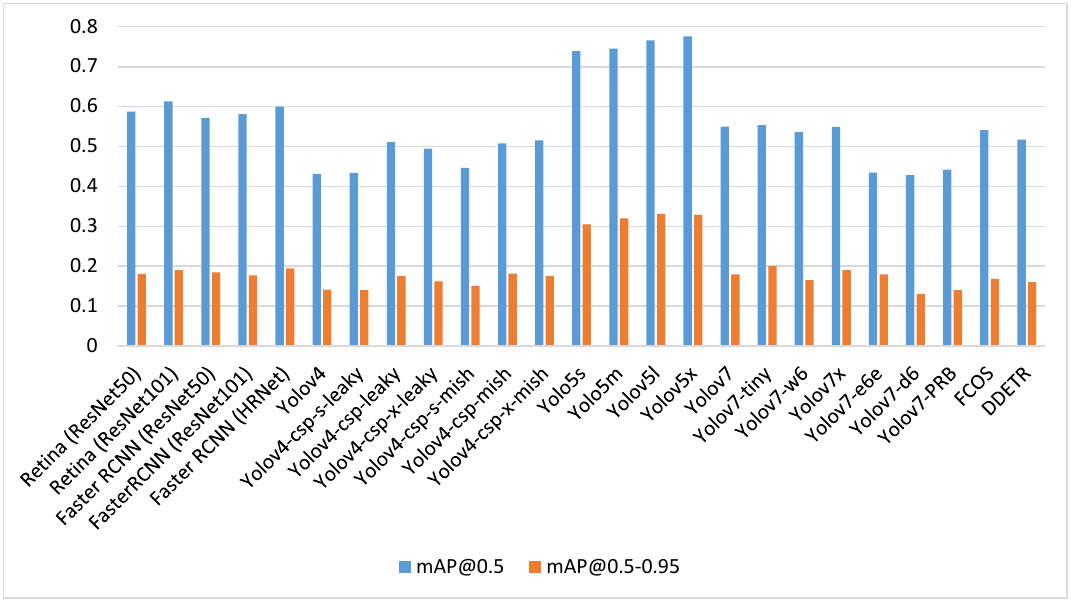}
    \caption{Detection accuracy of state-of-the-art baselines on our validation set.}
    \label{fig:mAP}
\end{figure}

We evaluated the detection performance of the baselines using the mean average precision (mAP) metric calculated by varying the detection scores of detected objects. True positives and false alarms were determined via the intersection over union (IOU) of detected objects and ground-truth objects. We calculated the mAP by thresholding the IOU using a threshold of 0.5 (mAP@0.5) and by averaging the mAP with thresholds varying in $[0.5, 0.95]$ (mAP@0.5-0.95).

\begin{figure}
    \centering
    \includegraphics[width=\columnwidth]{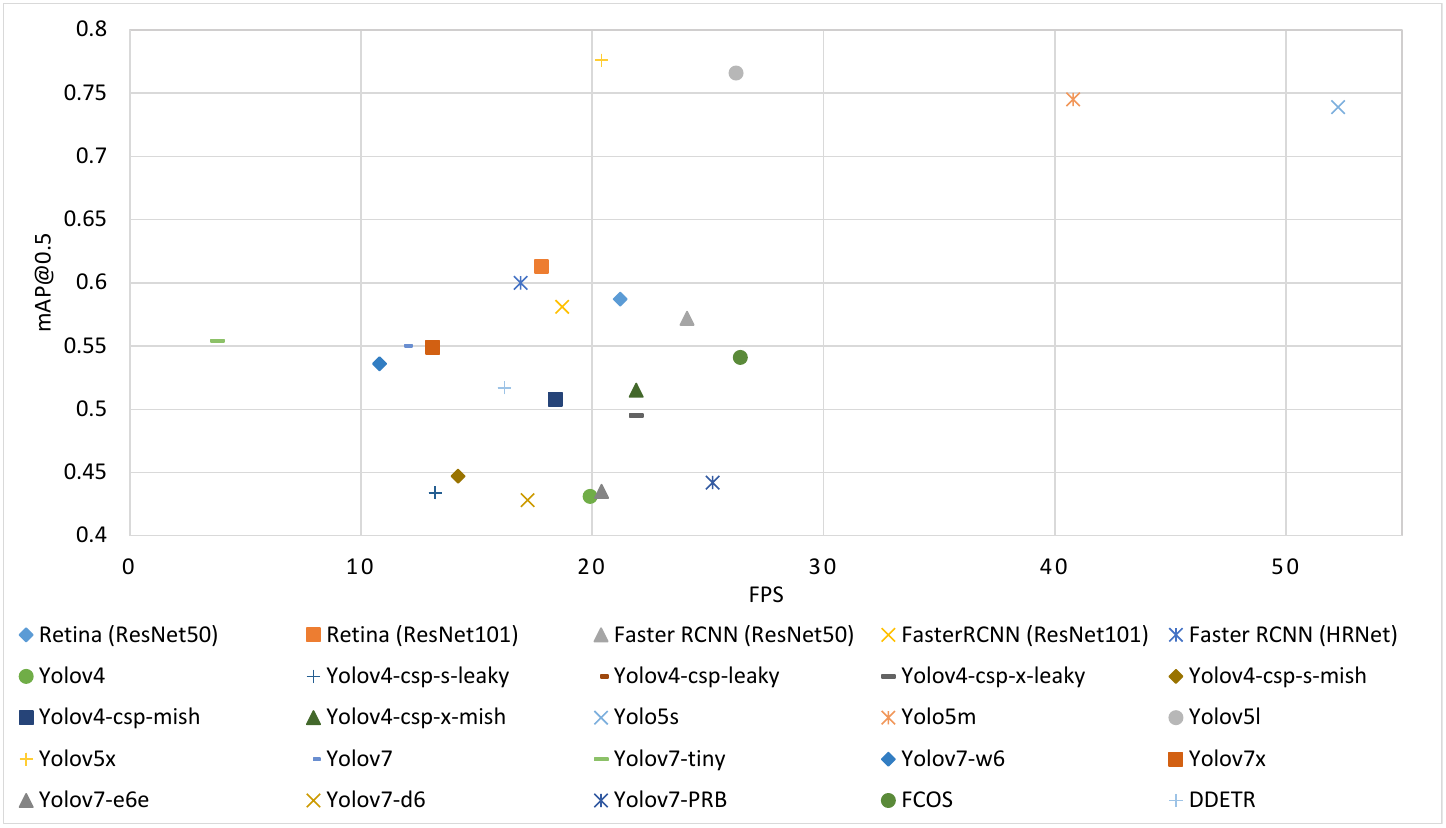}
    \caption{Detection accuracy vs. inference speed of state-of-the-art baselines on our validation set.}
    \label{fig:mAPvsFPS}
\end{figure}

\begin{figure*}
    \centering
    \includegraphics[width=0.49\textwidth]{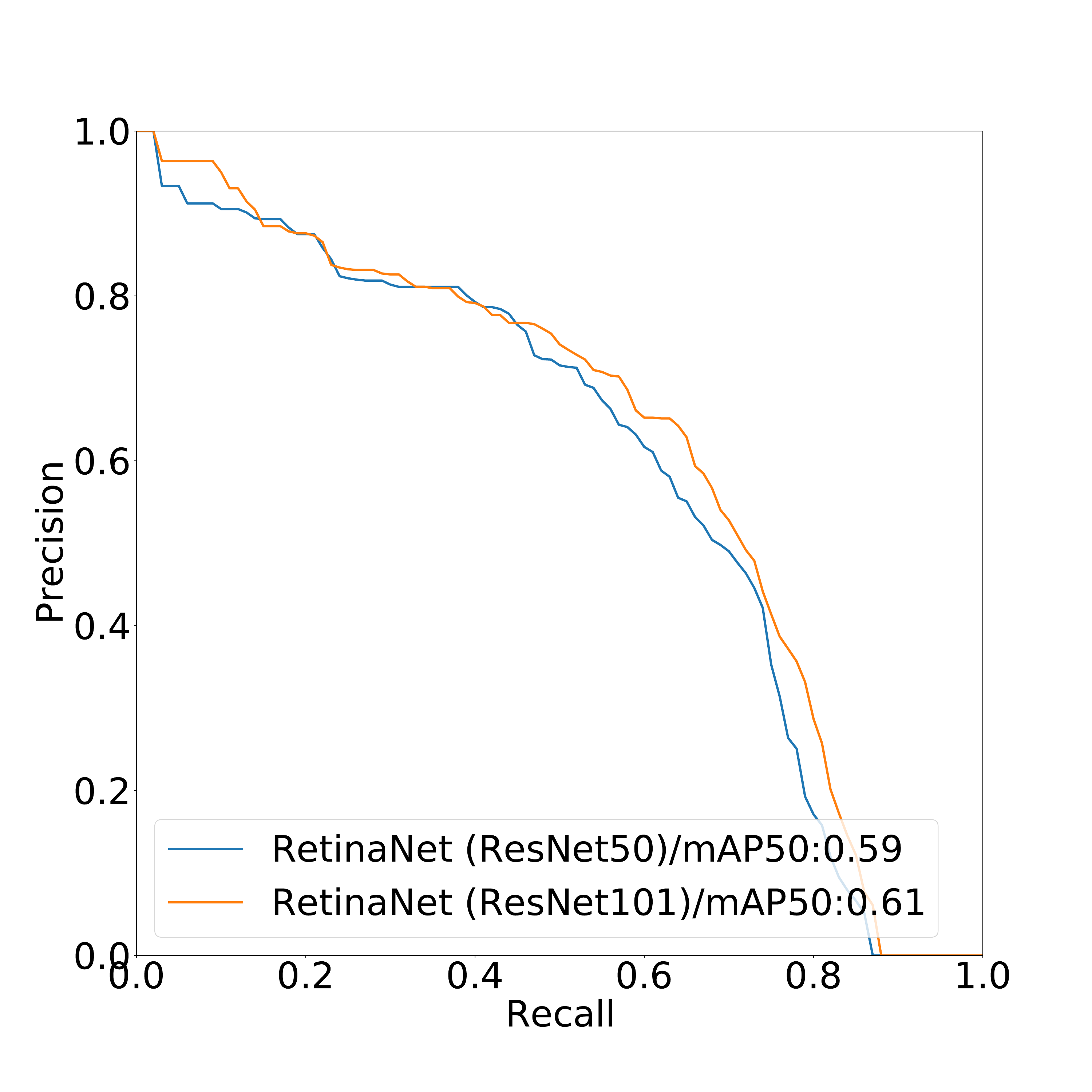} 
    \includegraphics[width=0.49\textwidth]
    {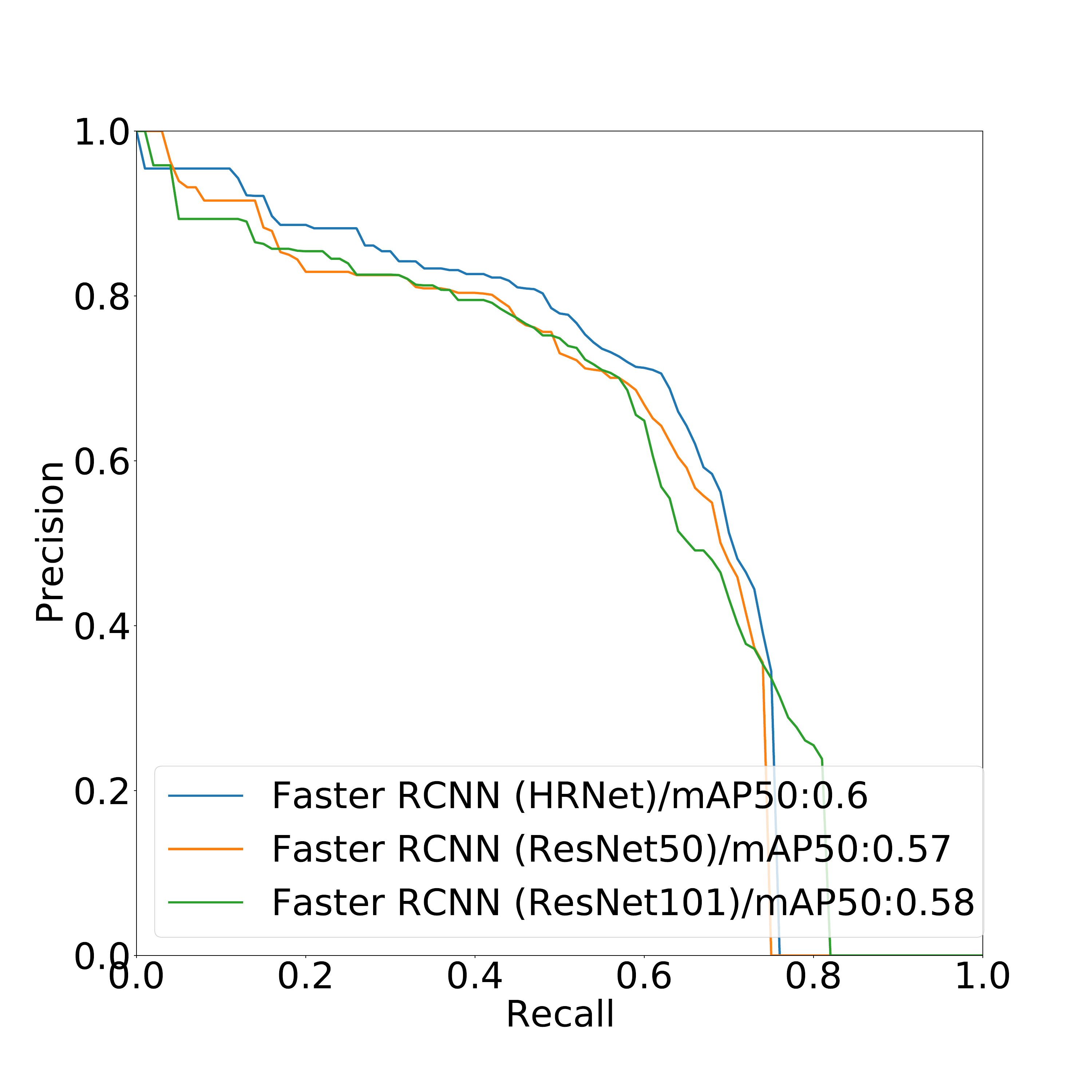} 
    \includegraphics[width=0.49\textwidth]{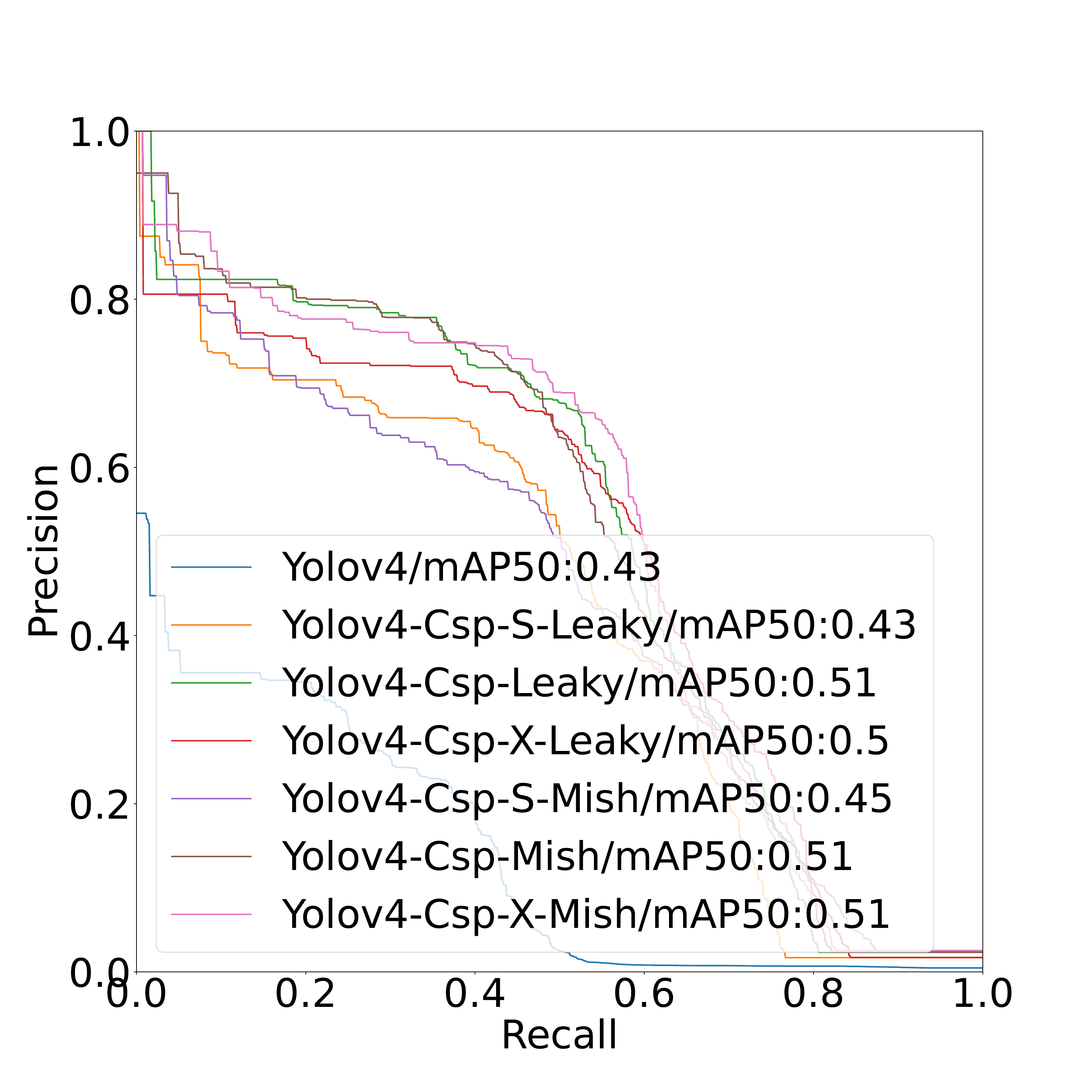} 
    \includegraphics[width=0.49\textwidth]{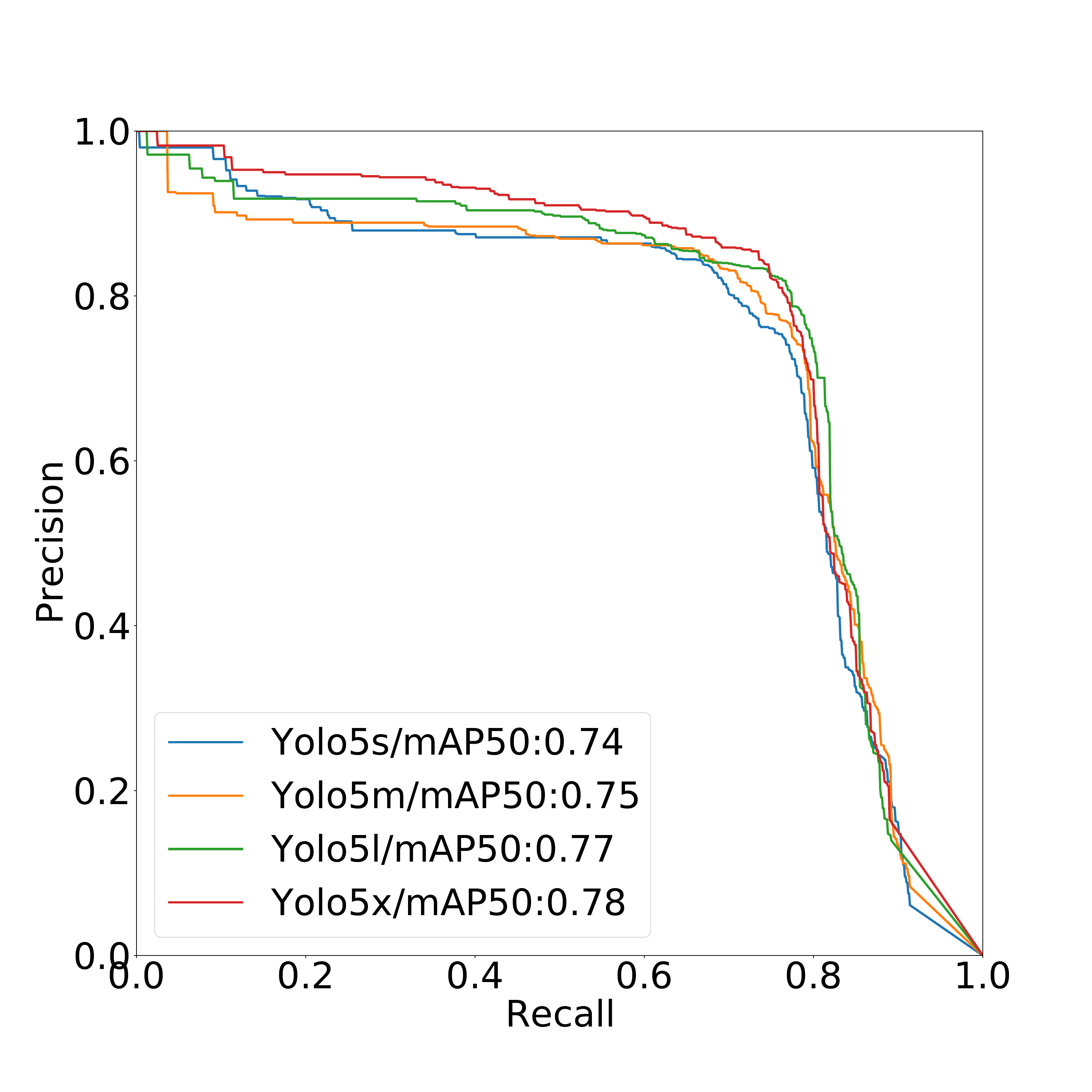} 
    \includegraphics[width=0.49\textwidth]{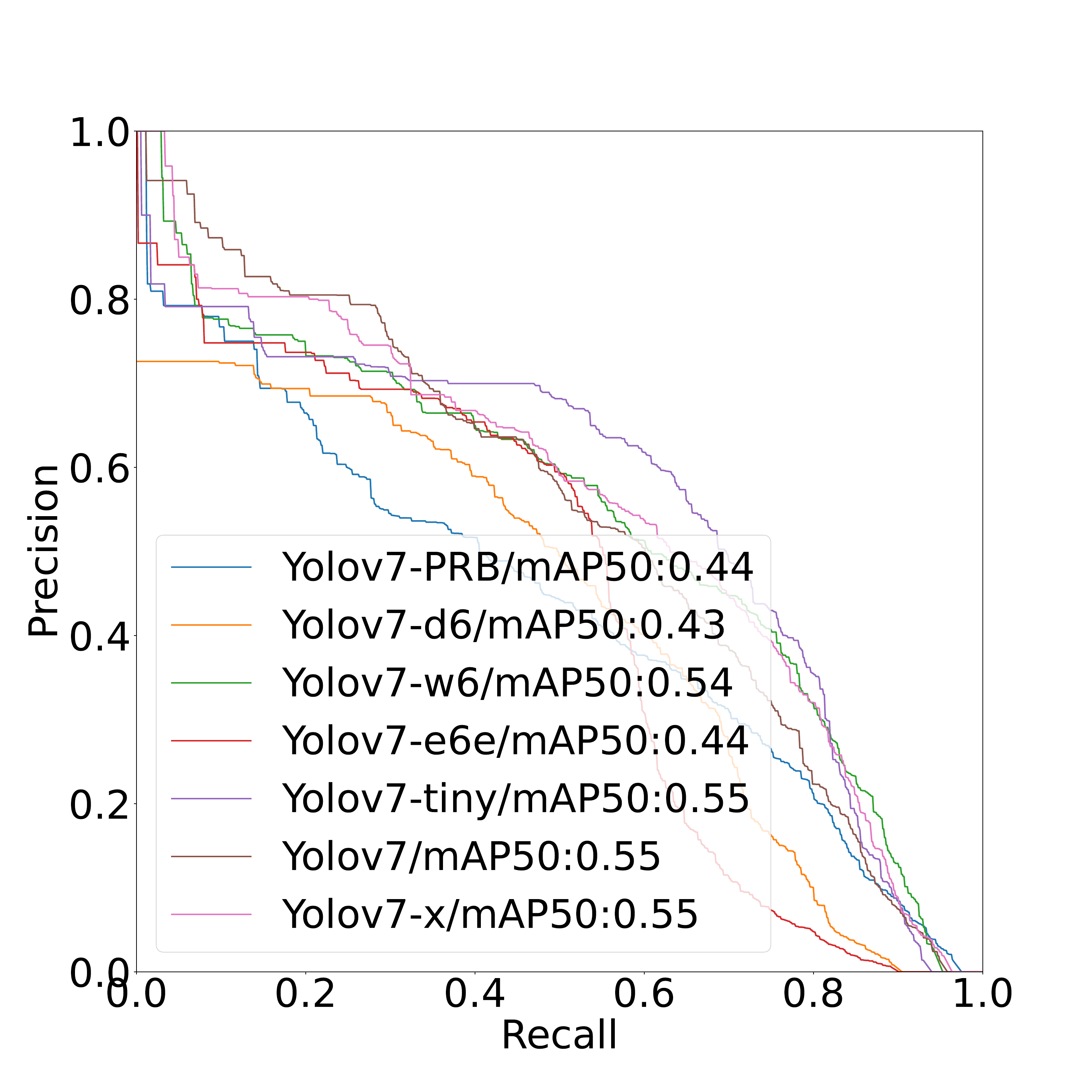} 
    \includegraphics[width=0.49\textwidth]{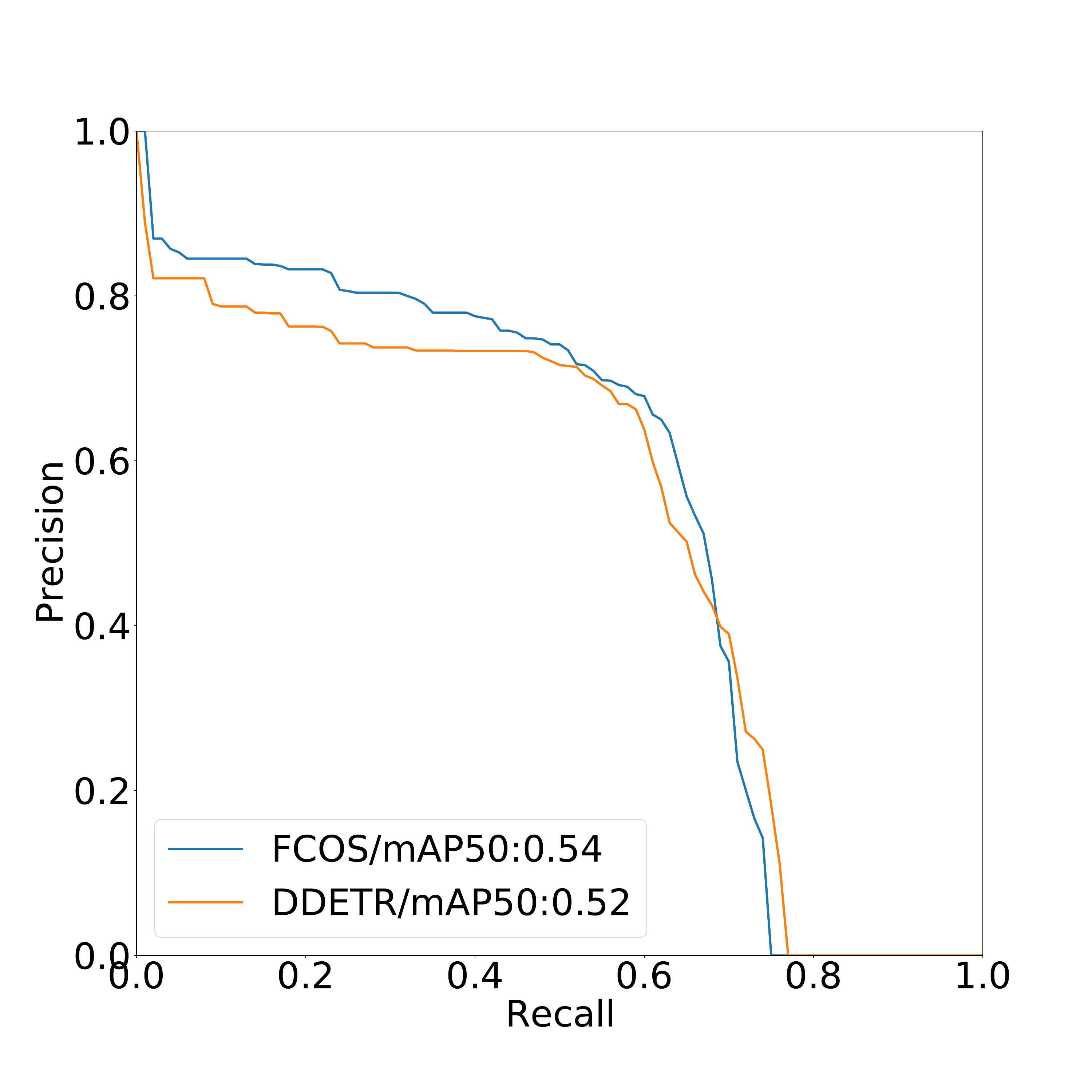} 
    \caption{PR curves of the baselines on our validation set.}
    \label{fig:PR}
\end{figure*}

We report the mAP@0.5 and mAP@0.5-0.95 of all the baselines and their variants on our validation set in Figure~\ref{fig:mAP}. We also show the detection accuracy (mAP@0.5) vs. the inference speed (fps) of all the experimented baselines in Figure~\ref{fig:mAPvsFPS}. To provide a deeper analysis into the performance of the baselines, we provide their precision-recall (PR) curves in Figure~\ref{fig:PR}, where relevant baselines are grouped into different graphs. 

As shown in experimental results (Figure~\ref{fig:mAP} and Figure~\ref{fig:mAPvsFPS}), the YOLO's family achieves superior performance over other baselines in both detection accuracy and computational efficiency. Its mAP@0.5 and mAP@0.5-0.95 are among the best, where YOLOv5x achieves the highest accuracy (0.78 mAP@0.5, 0.33 mAP@0.5-0.95), and performs reasonably fast (26.21 fps). We also observed that YOLOv5 is the best detector in the YOLO's family. YOLOv7 slightly outperforms YOLOv4 in terms of the detection accuracy while both the baselines have comparable performance in terms of the inference speed. In general, all variants of the YOLO's family clearly show the trade-off between model capacity and efficiency (i.e., larger models perform slower but get better accuracy). For instance, the tiny variant of YOLOv7 (YOLOv7-tiny) obtains the highest detection rate (0.55 mAP@0.5 and 0.20 mAP@0.5-0.95) among all the variants of the YOLOv7 baseline but also incurs the slowest speed (3.8 fps). Except for YOLOv7-tiny, both the YOLOv4's and YOLOv7's variants perform at around 10 to 25 fps.  

The RetinaNet baseline ranks second overall and its best architecture (with highest detection accuracy) is ResNet101. However, the difference in the detection accuracy between the ResNet50 and ResNet101 backbones is marginal (about 0.02 mAP@0.5 and 0.01 mAP@0.5-0.95). This baseline also shows relatively slow inference speed (21.2 fps and 17.8 fps for the ResNet50 and ResNet101 backbones respectively).

The Faster RCNN baseline seems to take the third place although the difference in its range of detection rates compared with that of the RetinaNet baseline is negligible. In particular, the range of detection rates of the Faster RCNN detector is $[0.572, 0.6]$ while it is $[0.587, 0.613]$ for the RetinaNet detector. Experimental results also show that the combination of the Faster RCNN baseline with the HRNet backbone outperforms other variants made of the ResNet50 and ResNet101 backbones. This is probably due to the design of the HRNet backbone to deal with high-resolution images. Note that, as shown in Table~\ref{tab:model_summary}, despite making use of less complex architecture (with less parameters and thus lower FLOPS), the Faster RCNN with the HRNet backbone takes longer inference time, compared with the RetinaNet with the ResNet101 backbone.

The two modern detectors, FCOS and DDETR, do not show advantages in our study. In addition, since those methods are anchor-free, they take time to learn bounding box locations but perform at the same speed (e.g., Faster RCNN) or slower than anchor-based baselines (e.g., YOLOv5).

\begin{figure*}
    \centering
    \includegraphics[width=0.49\textwidth]{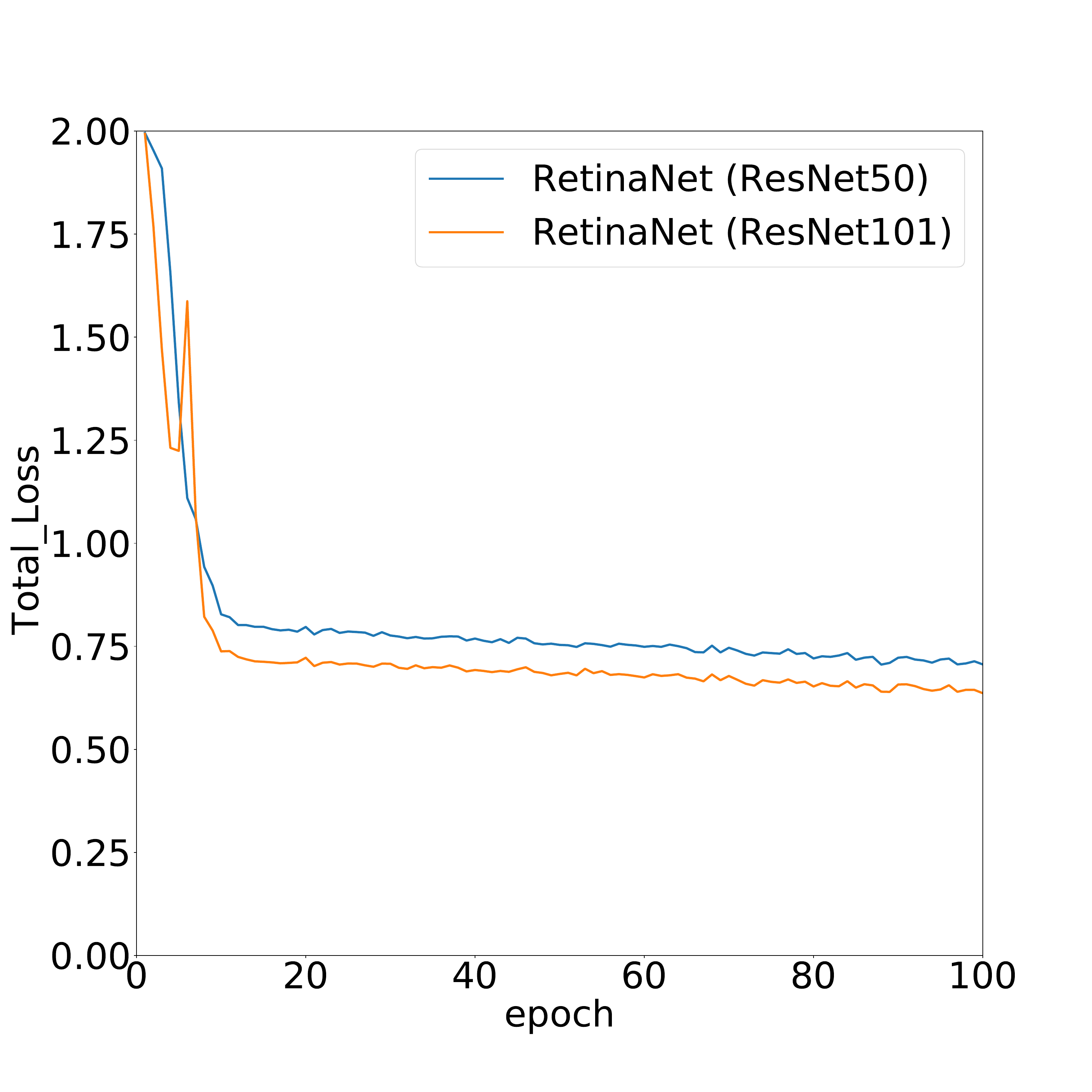} 
    \includegraphics[width=0.49\textwidth]{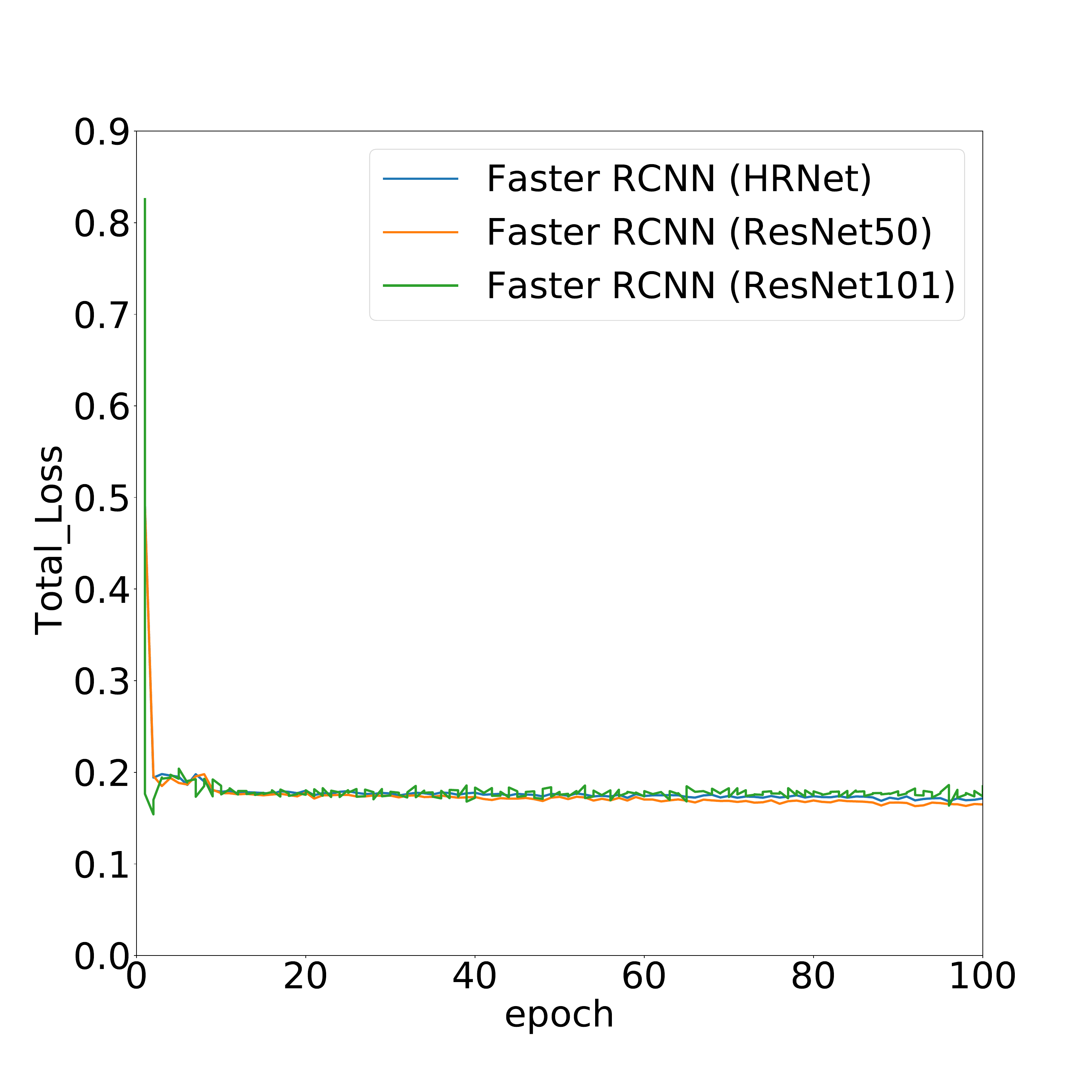} 
    \includegraphics[width=0.49\textwidth]{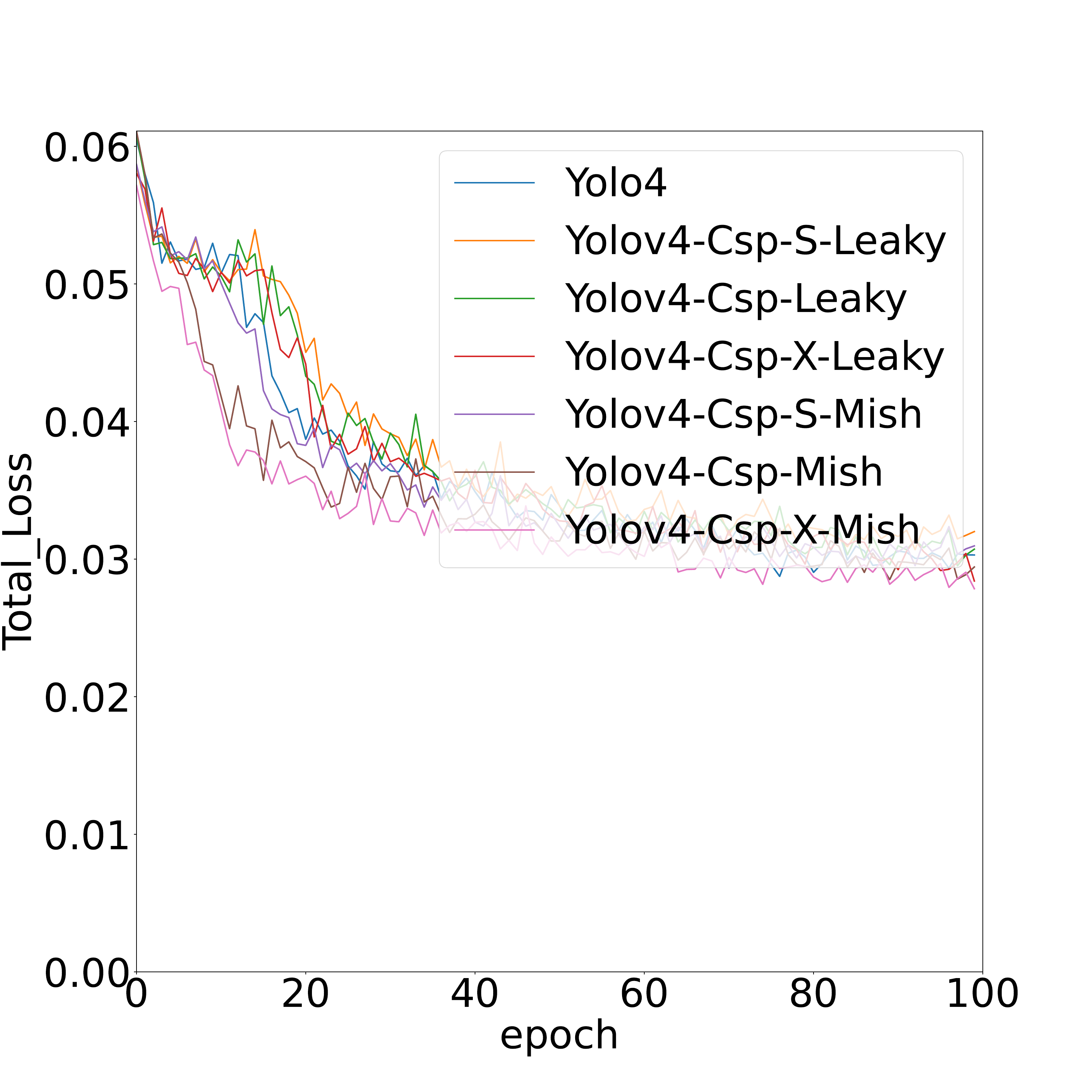} 
    \includegraphics[width=0.49\textwidth]{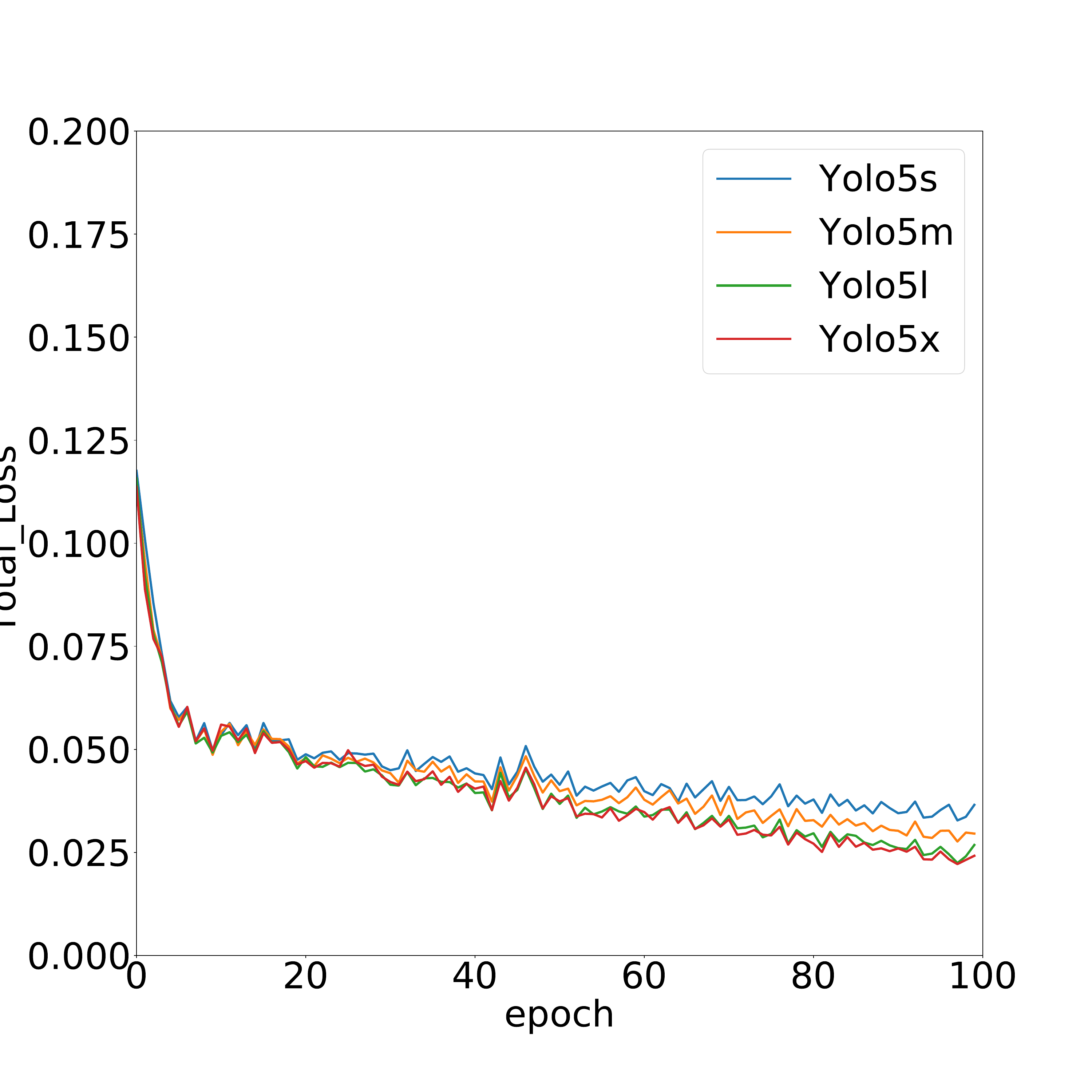} 
    \includegraphics[width=0.49\textwidth]{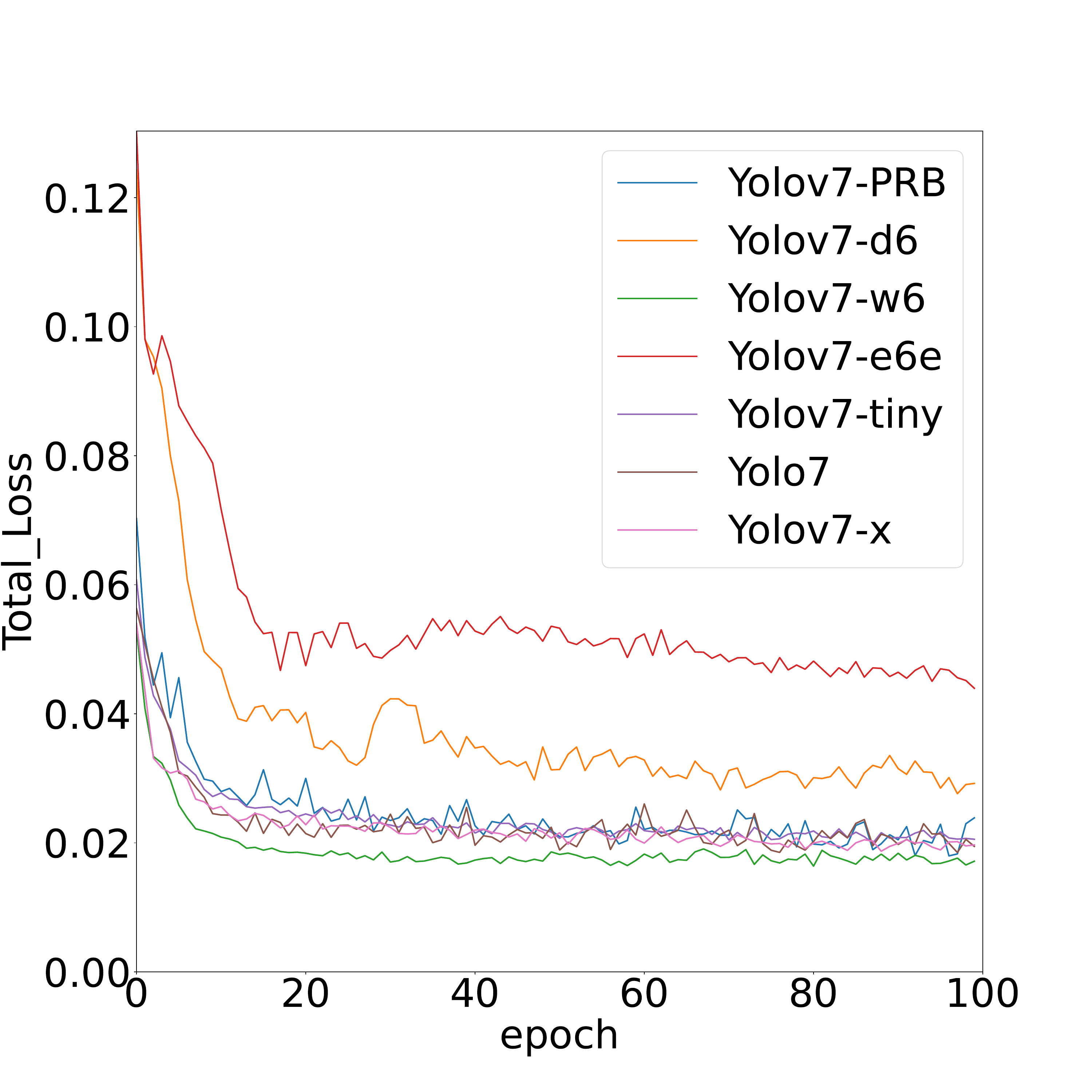} 
    \includegraphics[width=0.49\textwidth]{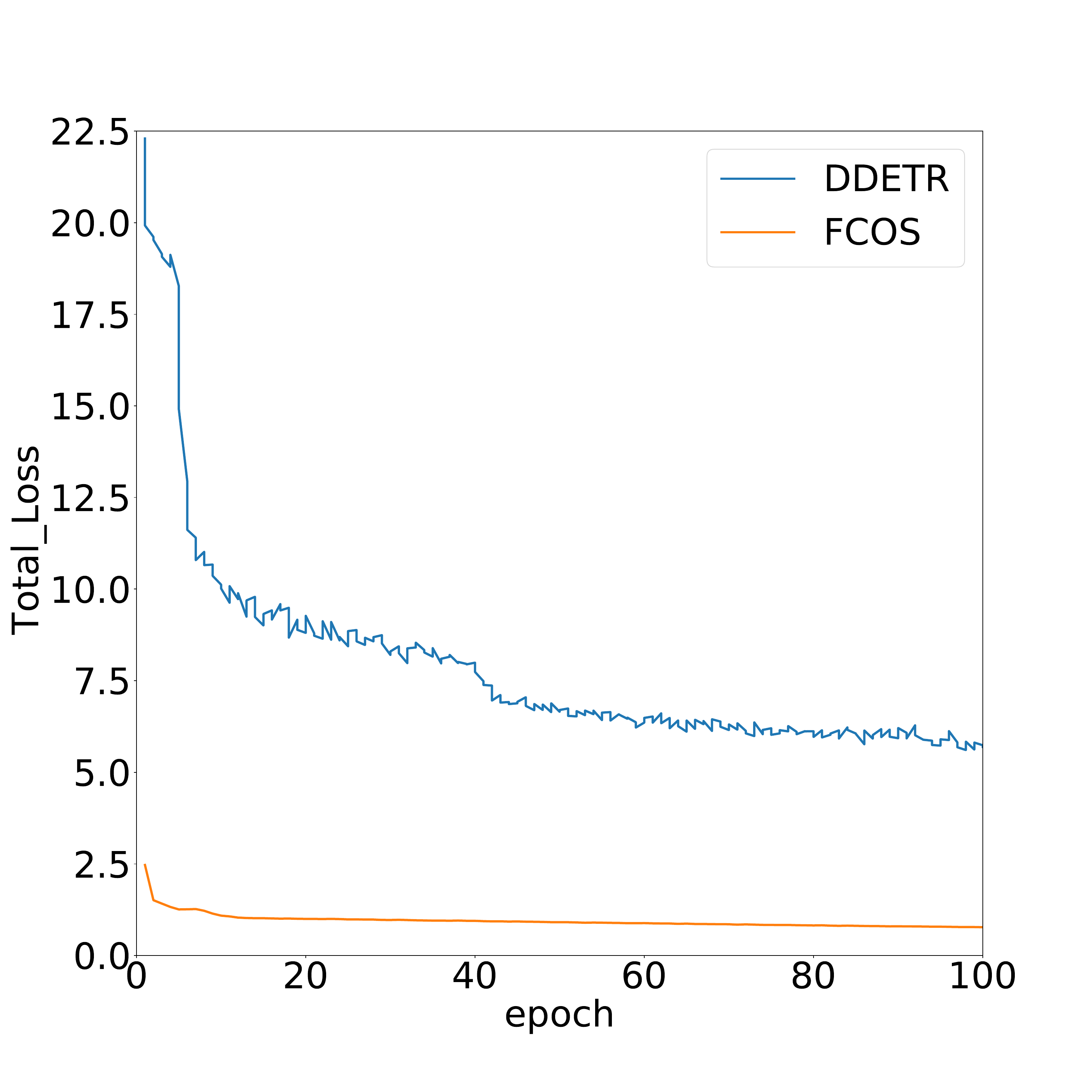} 
    \caption{Learning curves of the baselines on our training set.}
    \label{fig:learning_curves}
\end{figure*}

\begin{figure*}
    \centering
    \includegraphics[width=0.49\textwidth]{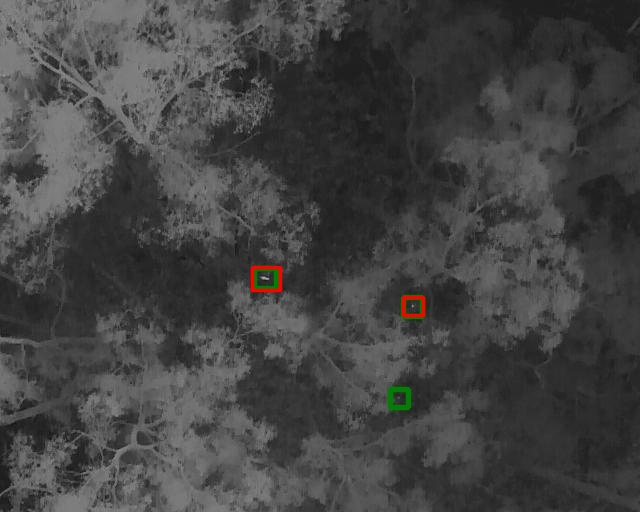}
    \includegraphics[width=0.49\textwidth]{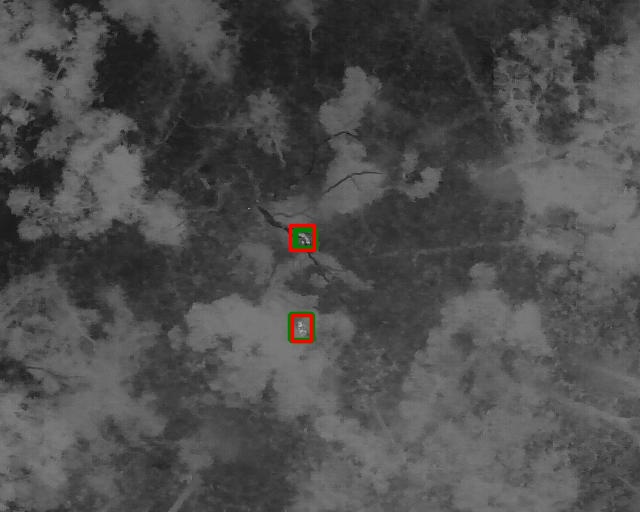}
    \includegraphics[width=0.49\textwidth]{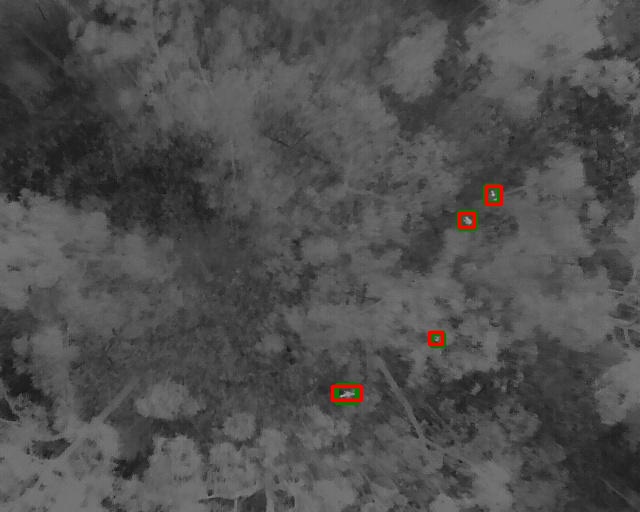}
    \includegraphics[width=0.49\textwidth]{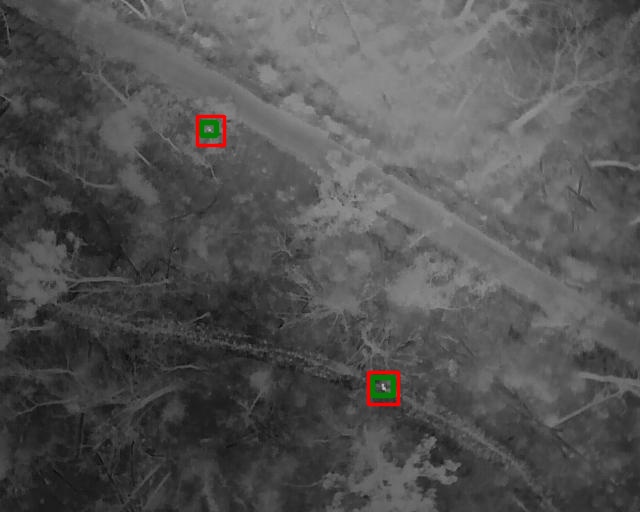}
    \includegraphics[width=0.49\textwidth]{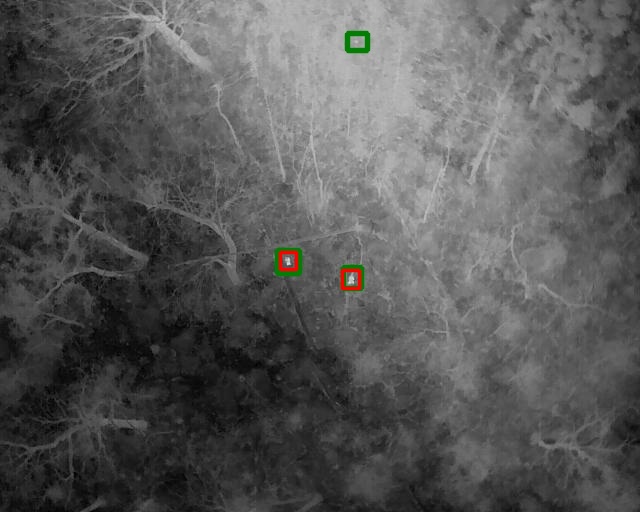}
    \includegraphics[width=0.49\textwidth]{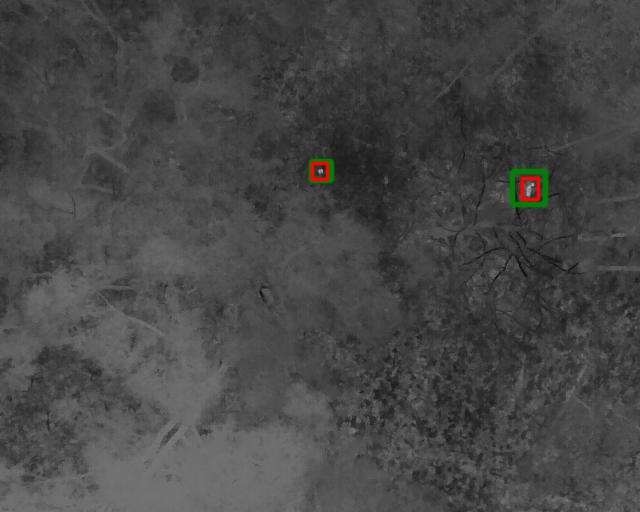}
    \includegraphics[width=0.49\textwidth]{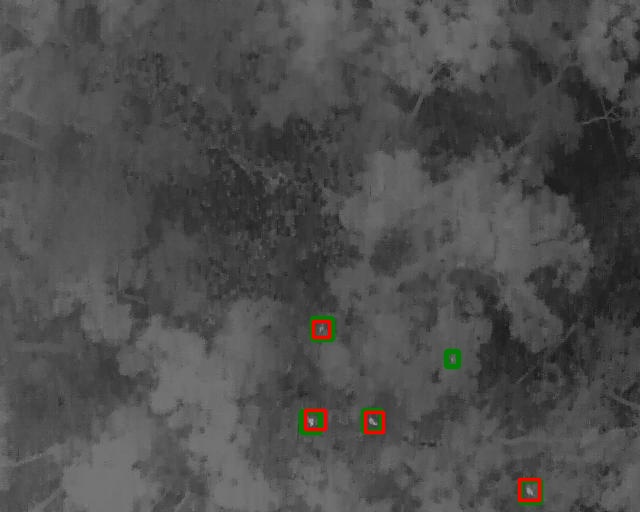}
    \includegraphics[width=0.49\textwidth]{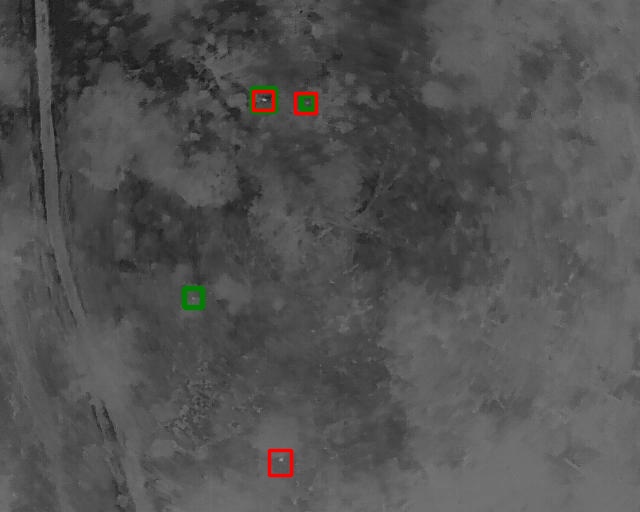}
    \caption{Several detection results of YOLOv5x. Detected objects and annotated objects are presented in green and red.}
    \label{fig:results}
\end{figure*}

To further investigate the learning process of the baselines, we plot the learning curves of all the baselines in Figure~\ref{fig:learning_curves}. We observed that all the baselines perform consistently on both the training and validation set. First, all the baselines show the convergence of their losses during the training, proving their learning ability. Second, the ranking of performance in both training and testing of all the baselines remains consistent. Specifically, as shown in Figure~\ref{fig:learning_curves}, YOLOv5 still appears as the best detector during training. It keeps improving during the training (the learning curves slide down continuously) and achieves the best error rate among all the baselines (around 0.025, shown in the vertical axis). In contrast, except for DDETR, other baselines quickly reach their stationary points. For instance, the learning curves of all the variants of Faster RCNN saturate at early epochs (around 10 epochs) with an error rate of 0.2. Likewise, RetinaNet follows this learning trend at early epochs, but then slightly improves. The learning of YOLOv4 and YOLOv7 is less stable, compared with the other baselines, evident by strong fluctuations in their loss curves. Both FCOS and DDETR, despite converging in the learning process, present high error rates. Specifically, the lowest error rates of both FCOS and DDETR are at several magnitude of those of the other baselines. DDETR incurs the highest error rate among all the baselines.

To showcase detection results, we choose YOLOv5x as it is the best detector and illustrate several results of this variant in Figure~\ref{fig:results}. In general, existing object detectors show their capability of detecting wildlife thermal signatures in realistic conditions. It remains challenging to identify animals presented in small size and in cluttered backgrounds. There requires a more effective design to upsample features in deep architectures to learn features from small-sized objects. We observed that, once detected wildlife species moved, they could be more easily detected by human annotators. This suggests the use of temporal information in improving animal detection.

\section{Conclusion}
\label{sec:conclusion}

This paper presents an empirical review of drone-based wildlife monitoring research since 2018. We provide a concise overview of existing methods regarding hardware specifications and settings, and object detection methods used in animal detection. To benchmark animal detection methods, we collected and annotated a real-world dataset of wildlife thermal signatures in a forested environment, and evaluated state-of-the-art object detectors on our dataset. Experimental results show that YOLOv5 significantly outperforms other baselines in both detection accuracy and computational speed. The benchmark results also suggest potential research directions including small-sized object detection and using temporal information in detecting animals from drone-derived imagery.

\section{Acknowledgments}

We acknowledge funding from CSIRO under the National Koala Monitoring Program and funding from the Victorian Government’s Department of Environment, Land, Water and Planning to conduct the surveys described in this study. We would also like to thank Desley Whisson for assistance selecting survey sites. We also thank Shelby Ryan for providing flight plans for the surveys described in this study. We also acknowledge Lachlan Clarke and Simon Ruff from the Victorian Government’s Department of Environment, Land, Water and Planning for facilitating access to our survey sites. We also acknowledge Blake Allan for assistance reviewing and facilitating these drone operations as Deakin University’s Chief Remote Pilot.

\bibliography{ref}

\end{document}